\documentclass[english]{article}
\usepackage[T1]{fontenc}
\usepackage[utf8]{inputenc}
\usepackage{float}
\usepackage{amsmath}
\usepackage{amssymb}
\usepackage{graphicx}
\usepackage[numbers]{natbib}

\makeatletter

\usepackage[final]{nips}

\usepackage{hyperref}       
\usepackage{url}            
\usepackage{booktabs}       
\usepackage{amsfonts}       
\usepackage{nicefrac}       
\usepackage{microtype}      

\newcommand{\comment}[1]{}

\makeatother

\usepackage{babel}
\begin{document}

\title{Modeling Friends and Foes}

\author{
Pedro A. Ortega\\
  DeepMind\\
  \texttt{pedroortega@google.com} \And
Shane Legg\\
  DeepMind\\
  \texttt{legg@google.com}
}

\maketitle

\begin{abstract}
How can one detect friendly and adversarial behavior from raw data?
Detecting whether an environment is a friend, a foe, or anything in
between, remains a poorly understood yet desirable ability for safe
and robust agents. This paper proposes a definition of these environmental
``attitudes'' based on an characterization of the environment's ability to 
react to the agent's private strategy. We define an objective function for a 
one-shot game that allows deriving the environment's probability distribution 
under friendly and adversarial assumptions alongside the agent's optimal 
strategy. Furthermore, we present an algorithm to compute these equilibrium 
strategies, and show experimentally that both friendly and adversarial 
environments possess non-trivial optimal strategies.
\end{abstract}

\begin{quote}
\end{quote}

\section{Introduction}

How can agents detect friendly and adversarial behavior from raw data?
Discovering whether an environment (or a part within) is a friend or a foe 
is a poorly understood yet desirable skill for safe and robust agents.
Possessing this skill is important for a number of situations, including:
\begin{itemize}
\item \emph{Multi-agent systems}: Some environments, especially in 
multi-agent systems, might have incentives to either help or hinder the agent 
\citep{Leike2017}. For example, an agent playing football must anticipate 
both the creative moves of its team members and its opponents. Thus, learning 
to discern between friends and foes might not only help the agent to avoid 
danger, but also open the possibility to solving taks throught collaboration 
that it could not solve alone otherwise.
\item \emph{Model uncertainty}: An agent can choose to impute 
``adversarial'' or ``friendly'' qualities to an environment that it does not 
know well. For instance, an agent that is trained in a simulator could 
compensate for the innaccuracies by assuming that the real environment differs 
from the simulated one---but in an adversarial way, so as to devise  
countermeasures ahead of time~\citep{Amodei2016}. Similarly, innacuracies 
might also originate in the agent itself---for instance, due to bounded 
rationality \citep{Russell1997, 
OrtegaBraun2013}.
\end{itemize}
Typically, these situations involve a \emph{knowledge limitation} 
that the agent addresses by responding with a \emph{risk-sensitive} policy.

The contributions of this paper are threefold. First, we offer a 
\emph{broad definition of friendly and adversarial behavior}. Furthermore, by 
varying a single real-valued parameter, one can select from a continuous range 
of behaviors that smoothly interpolate between fully adversarial and fully 
friendly. Second, we derive the agent's (and environment's) \emph{optimal 
strategy} under friendly or adversarial assumptions. To do so, we treat the 
agent-environment interaction as a one-shot game with information-constraints,
and characterize the optimal strategies at equilibrium. Finally, we
provide an \emph{algorithm to find the equilibrium strategies} of the agent and 
the environment. We also demonstrate empirically that the resulting strategies 
display non-trivial behavior which vary qualitatively with the 
information-constraints.

\section{Motivation \& Intuition}\label{sec:rationale}

We begin with an intuitive example using multi-armed 
bandits. This helps motivating our mathematical formalization in the 
next section.

Game theory is the classical economic paradigm to analyze the interaction
between agents~\citep{OsborneRubinstein1994}. However, within game
theory, the term \emph{adversary} is justified by the fact that in
zero-sum games the equilibrium strategies are \emph{maximin} strategies,
that is, strategies that maximize the expected payoff under the assumption
that the adversary will minimize the payoffs. However, when the game
is not a zero-sum game, interpreting an agent's behavior as adversarial
is far less obvious, as there are is no coupling between 
payoffs, and the strong guarantees provided by the minimax theorem are 
unavailable \cite{VonNeumann1947}. The notions of ``indifferent'' and 
``friendly'' are similarly troublesome to capture using the standard 
game-theoretic language.

\begin{figure}[H]
\centering{} \includegraphics[width=0.8\textwidth]{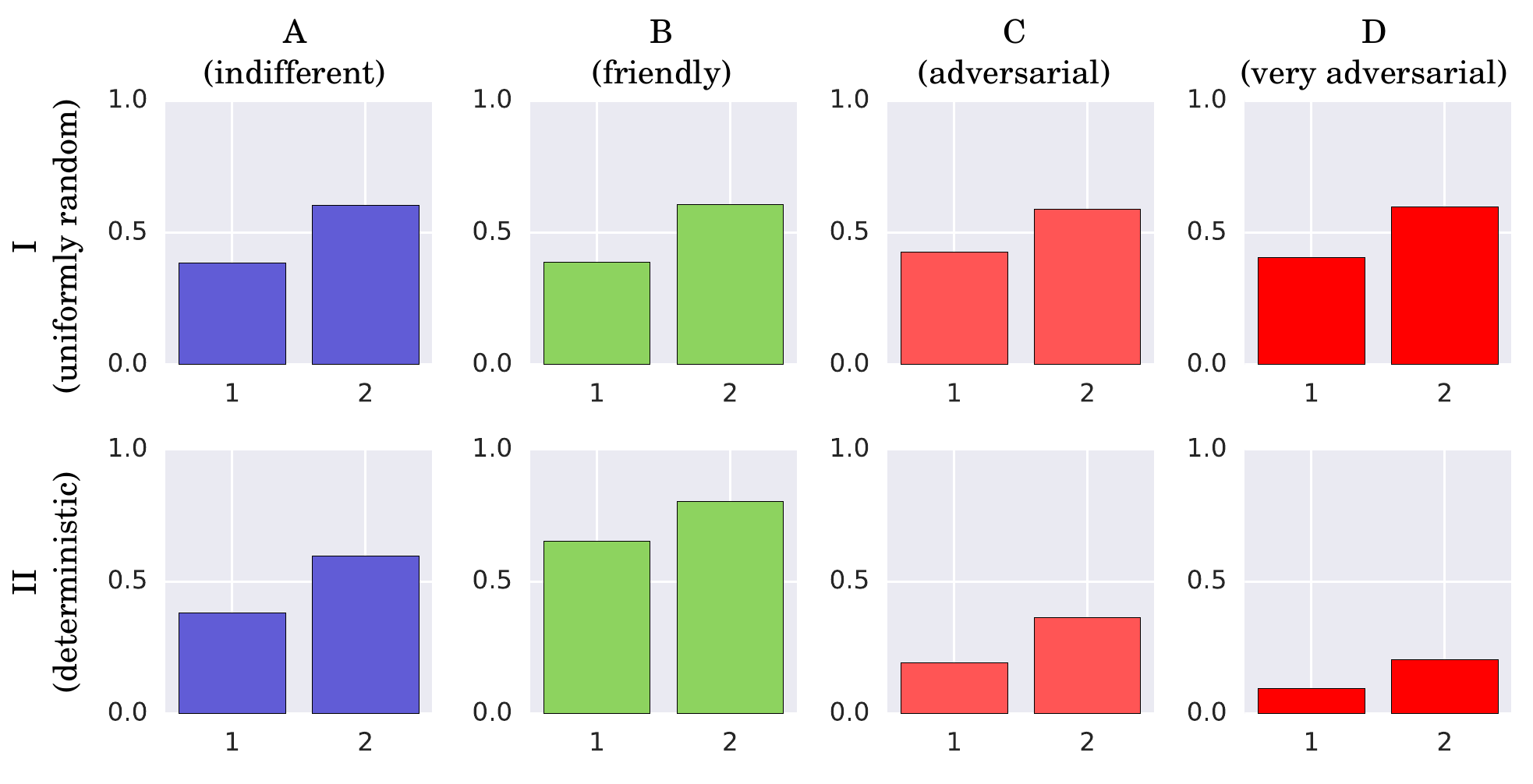}
\caption{Average rewards for four different \emph{two-armed bandits} (columns 
A--D) under two different strategies: (I) a uniform strategy (arms were
chosen using 1000 fair coin flips) and (II) a deterministic strategy
(500 times arm~1, then 500 times arm~2). Bandits get to choose in each turn 
(using a probabilistic rule) which one of the two arms will deliver the 
reward. The precise probabilistic rule used here will be explained later 
in Section~\ref{sec:experiments} (Experiments).\label{fig:bernoulli-mods}}
\end{figure}

Intuitively though, terms such as \emph{adversarial}, \emph{indifferent},
and \emph{friendly} have simple meanings. To illustrate, consider
two different strategies (I~\&~II) used on four different two-armed bandits
(named A--D), where each strategy gets to play 1000 rounds. In each
round, a bandit secretly chooses which one (and only one) of the two arms will 
deliver the reward. Importantly, each bandit chooses the location of 
the arm using a different probabilistic rule. Then the agent pulls an arm, 
receiving the reward if it guesses correctly and nothing otherwise. 

The two different strategies are:
\begin{description}
\item [{I}] The agent plays all 1000 rounds using a \emph{uniformly random} 
strategy. 
\item [{II}] The agent \emph{deterministically} pulls each arm exactly
500 times using some fixed rule.
\end{description}
Note that each strategy pulls each arm approximately 50\% of the time.
Now, the agent's average rewards for all four bandits are shown in 
Figure~\ref{fig:bernoulli-mods}.
Based on these results, we make the following observations:
\begin{description}
\item[1.] \emph{Sensitivity to strategy}. The two sets of average rewards for 
bandit~A are statistically indistinguishable, that is, they stay the same 
regardless of the agent's strategy. This 
corresponds to the \emph{stochastic bandit} type in the 
literature \citep{SuttonBarto1998, BubeckCesaBianchi2012}. In contrast, 
bandits~B--D yielded \emph{different} average rewards for the two strategies. 
Although each arm was pulled approximately 500 times, \emph{it appears as if} 
the reward distributions were a function of the strategy. 

\item[2.] \emph{Adversarial/friendly exploitation of strategy}. The average 
rewards 
do not always add up to one, as one would expect if the rewards were truly 
independent of the strategy. Compared to the uniform strategy, the 
deterministic strategy led to either an
increase (bandit~B) or a decrease of the empirical rewards (bandits~C
and~D). We can interpret the behavior of bandits C~\&~D as an adversarial
exploitation of the predictability of the agent's strategy---much
like an exploitation of a deterministic strategy in a rock-paper-scissors
game. Analogously, bandit~B appears to be friendly towards the agent,
tending to place the rewards favorably.

\item[3.] \emph{Strength of exploitation}. Notice how the rewards of
both adversarial bandits (C~\&~D) when using strategy~II differ
in how strongly they deviate from the baseline set by strategy~I.
This difference suggests that bandit~D is better at reacting to the
agent's strategy than bandit~C---and therefore also more adversarial. A bandit 
that can freely choose \emph{any} placement of rewards is known as a 
non-stochastic bandit \citep{AuerEtAl2002, BubeckCesaBianchi2012}.

\item[4.] \emph{Cooperating/hedging}. The nature of the 
bandit qualitatively affects the agent's optimal strategy. A friendly bandit 
(B) invites the agent to cooperate through the use of predictable policies; 
whereas adversarial bandits (C \& D) pressure the agent to hedge through 
randomization. \end{description}

\begin{figure}[H]
\centering{} \includegraphics[width=0.5\textwidth]{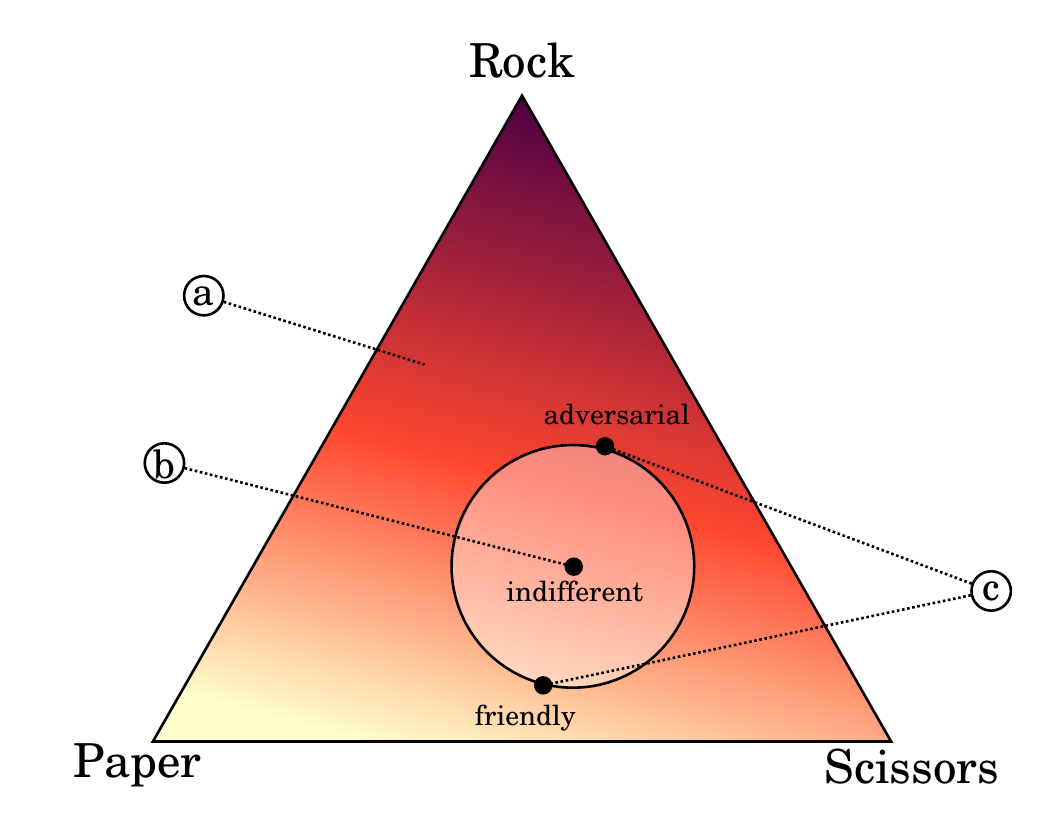}
\caption{Reacting to the agent's strategy in Rock-Paper-Scissors. The diagram 
depicts the simplex of the \emph{environment's mixed strategies} over the three 
pure strategies Rock, Paper, and Scissors, located at the corners. a) When the 
agent picks a strategy it fixes its expected payoff (shown in color, 
where darker is worse and lighter is better). b) An indifferent environment 
corresponds to a player using a fixed strategy (mixed or pure). c) However, a 
reactive environment can deviate from the indifferent strategy (the set of 
choices is shown as a ball). A friendly environment would choose in a way that 
benefits the player (i.e.\ try playing Paper as much as possible if the player 
mostly plays Scissors); analogously, an adversarial environment would attempt to 
play the worst strategy for the agent.\label{fig:channel}}
\end{figure}

Simply put, bandits B--D appear to react to the agent's \emph{private
strategy} in order to manipulate the payoffs. Abstractly, we can
picture this as follows. First, a reactive environment can be thought
of as possessing privileged information about the agent's private strategy, 
acquired e.g.\ through past experience or through ``spying''. Then, the 
amount of information determines the extent to which the environment 
is willing to deviate from a baseline, indifferent strategy. Second, the 
adversarial or friendly nature of the environment is reflected by the strategy 
it chooses: an adversarial (resp.~friendly) environment will select 
the reaction that minimizes (resp.~maximizes) the agent's payoff. The diagram 
in Figure~\ref{fig:channel} illustrates this idea in a Rock-Paper-Scissors game.

We make one final observation.
\begin{description}
 \item[5.] \emph{Agent/environment symmetry}. Let us turn the tables on the 
agent: how should we play if we were the bandit? A moment of reflection 
reveals that the analysis is symmetrical. An agent that does not attempt to 
maximize the payoff, or cannot do so due to limited reasoning power, will 
pick its strategy in a way that is indifferent to our placement of the reward. 
In contrast, a more effective agent will react to our choice, seemingly 
anticipating it. Furthermore, the agent will appear friendly if our goal is to 
maximize the payoff and adversarial if our goal is to minimize it.
\end{description}
This symmetry implies that the choices of the agent and the environment 
are coupled to each other, suggesting a solution principle for determining a 
strategy profile akin to a Nash equilibrium \citep{OsborneRubinstein1994}. The 
next section will provide a concrete formalization. 

\section{Characterizing Friends and Foes}\label{sec:formalization}

In this section we formalize the picture sketched out in the preceeding
section. We first state an objective function for a game that couples
the agent's interests with those of the environment and limits both
player's ability to react to each other. We then derive expressions
for their optimal strategies (i.e.~the best-response functions).
Based on these, we then give an existence plus an indifference result
for the ensuing equilibrium strategies.

\subsection{Objective function}

Let $\mathcal{X}$ and $\mathcal{Z}$ denote the set of actions (i.e.\ pure 
strategies) of the agent and the environment respectively; and let $Q$ and 
$P$ denote prior and posterior strategies respectively. We represent the 
interaction between the agent and the environment as a \emph{one-shot game} in
which both the agent, starting from prior strategies 
$Q(X)\in\Delta(\mathcal{X})$ and $Q(Z)\in\Delta(\mathcal{Z})$, choose (mixed) 
posterior strategies $P(X)\in\Delta(\mathcal{X})$ and 
$P(Z)\in\Delta(\mathcal{Z})$ respectively.
The goal of the agent is to maximize the payoffs given by a function
that maps choices $(x,z)\in\mathcal{X}\times\mathcal{Z}$ into utilities
$U(x,z)\in\mathbb{R}$. 

We model the two players' sensitivity to each other's strategy as
coupled deviations from indifferent prior strategies, whereby each
player attempts to extremize the expected utility, possibly pulling
in opposite directions. Formally, consider the objective function

\begin{equation}
\begin{matrix}
 J & = & \mathbb{E}[U(X,Z)] 
   & - & \frac{1}{\alpha}D_\mathrm{KL}(P(X)\|Q(X))
   & - & \frac{1}{\beta}D_\mathrm{KL}(P(Z)\|Q(Z)) \\ \\
   & = & \sum_{x,z}P(x)P(z)U(x,z) 
   & - & \frac{1}{\alpha}\sum_{x}P(x)\log\frac{P(x)}{Q(x)}
   & - & \frac{1}{\beta}\sum_{z}P(z)\log\frac{P(z)}{Q(z)} \\ \\
   & = & \text{\{coupled expected payoffs\}}
   & - & \text{\{agent deviation cost\}}
   & - & \text{\{env. deviation cost\}}\\
\end{matrix}\label{eq:objective}
\end{equation}
where $\alpha,\beta\in\mathbb{R}$, known as the \emph{inverse temperature} 
parameters, determine the reaction abilities of the agent and the 
environment respectively. 

This objective function~(\ref{eq:objective})
is obtained by coupling two free energy functionals (one for each
player) which model decision-making with information-constraints (see
e.g.\ \citep{TishbyPolani2011,OrtegaBraun2013}). We will discuss the 
interpretation of this choice further in 
Section~\ref{sec:information-channel}. Other constraints are possible, e.g.\ 
any deviation quantified as a Bregman divergence \citep{Banerjee2005}.

\subsection{Friendly and Adversarial Environments}

Both the sign and the magnitude of the inverse temperatures control the 
player's reactions as follows. 
\begin{description}
\item[1.] The \emph{sign} of
$\beta$ determines the extremum operation: if $\beta$ is positive,
then $J$ is concave for fixed $P(X)$ and the environment maximizes
the objective w.r.t.~$P(Z)$; analogously, a negative value of $\beta$
yields a convex objective~$J$ that is minimized w.r.t~$P(Z)$.
\item[2.] The \emph{magnitude} of $\beta$ determines the strength of the 
deviation:
when $|\beta|\approx0$, the environment can only pick strategies
$P(Z)$ that are within a small neighborhood of the center~$Q(Z)$,
whereas $|\beta|\gg0$ yields a richer set of choices for $P(Z)$.
\end{description}
The parameter $\alpha$ plays an analogous role, although in this
exposition we will focus on $\alpha\geq0$ and interpret it as a parameter
that controls the agent's ability to react. In particular, setting
$\alpha=0$ fixes $P(X)$ to $Q(X)$, which is useful for deriving the 
posterior environment $P(Z)$ for a given, fixed agent strategy. 

From the above, it is easy to see that friendly and adversarial environments
are modeled through the appropriate choice of $\beta$. For $\alpha>0$
and $\beta>0$, we obtain a friendly environment that helps the agent
in maximizing the objective, i.e.~
\[
\max_{P(X)}\max_{P(Z)}\{J[P(X),P(Z)]\}.
\]
In contrast, for $\alpha>0$ and $\beta<0$, we get an adversarial
environment that minimizes the objective:
\[
\max_{P(X)}\min_{P(Z)}\{J[P(X),P(Z)]\}=\min_{P(Z)}\max_{P(X)}\{J[P(X),P(Z)]\}.
\]
In particular, the equality after exchanging the order of the minimization and 
maximization can be shown to hold using the
\emph{minimax theorem}: $J$ is a continuous and convex-concave function
of $P(X)$ and $P(Z)$, which in turn live in compact and convex sets
$\Delta(\mathcal{X})$ and $\Delta(\mathcal{Z})$ respectively. The
resulting strategies then locate a saddle point of $J$.

\subsection{Existence and Characterization of Equilibria}

To find the equilibrium strategies for (\ref{eq:objective}), we calculate
the \emph{best-response function} for each player, i.e.~the optimal
strategy for a given strategy of the other player in both the friendly and 
adversarial cases. Proofs to the claims can be found in 
Appendix~\ref{sec:proofs}.

\paragraph{Proposition 1. }

The best-response functions $f_{X}, f_{Z}$ for the agent and the environment
respectively are given by the Gibbs distributions
\begin{align}
P(X) & =f_{X}[P(Z)]: & P(x) & =\frac{1}{N_{X}}Q(x)\exp\{\alpha U(x)\}, & U(x) & :=\sum_{z}P(z)U(x,z);\label{eq:opt-player}\\
P(Z) & =f_{Z}[P(X)]: & P(z) & =\frac{1}{N_{Z}}Q(z)\exp\{\beta U(z)\}, & U(z) & :=\sum_{x}P(x)U(x,z)\label{eq:opt-env}
\end{align}
respectively, where $N_{X}$ and $N_{Z}$ are normalizing constants. $\square$

Given the above best-response functions $f_{X}$ and $f_{Z}$, we
define an \emph{equilibrium strategy profile} of the objective function
(\ref{eq:objective}) as a fixed-point of the \emph{combined best-response
function} defined as a mapping $f:\Delta(\mathcal{X})\times\Delta(\mathcal{Z})\rightarrow\Delta(\mathcal{X})\times\Delta(\mathcal{Z})$
that concatenates the two best-response functions, i.e.
\begin{equation}
f[P(X),P(Z)]:=(f_{X}[P(Z)],f_{Z}[P(X)]).\label{eq:comb-best-resp}
\end{equation}
That is, the equilibrium strategy profile%
\footnote{This definition is closely related to the Quantal-Response-Equilibrium
\citep{McKelveyPalfrey1995}.%
}, in analogy with the Nash equilibrium, is a mixed-strategy profile 
that lies at the intersection of both best-response curves. With this 
definition, the following existence result follows immediately.

\paragraph{Proposition 2.}
There always exists an equilibrium strategy profile. $\square$

Finally, the following result characterizes the equilibrium strategy
profile in terms of an indifference principle (later illustrated in 
Figure~\ref{fig:learning-dynamics}). In particular, the
result shows that both players strive towards playing strategies that
equalize each other's \emph{net payoffs} defined as
\begin{equation}
J_{X}(x) := \alpha\sum_{z}P(z)U(x,z) - \log\frac{P(x)}{Q(x)}
\quad \text{and} \quad 
J_{Z}(z) := \beta\sum_{x}P(x)U(x,z) - \log\frac{P(z)}{Q(z)}.
\label{eq:net-payoffs}
\end{equation}

\paragraph{Proposition 3.}

In equilibrium, the net payoffs are such that for all $x,x'\in\mathcal{X}$
and all $z,z'\in\mathcal{Z}$ in the support of $P(X)$ and $P(Z)$
respectively, 
\[
J_{X}(x)=J_{X}(x')\quad\text{and}\quad J_{Z}(z)=J_{Z}(z'). \square
\]

\section{Computing Equilibria\label{sec:learning}}

\begin{figure}[H]
\begin{centering}
\includegraphics[width=1\textwidth]{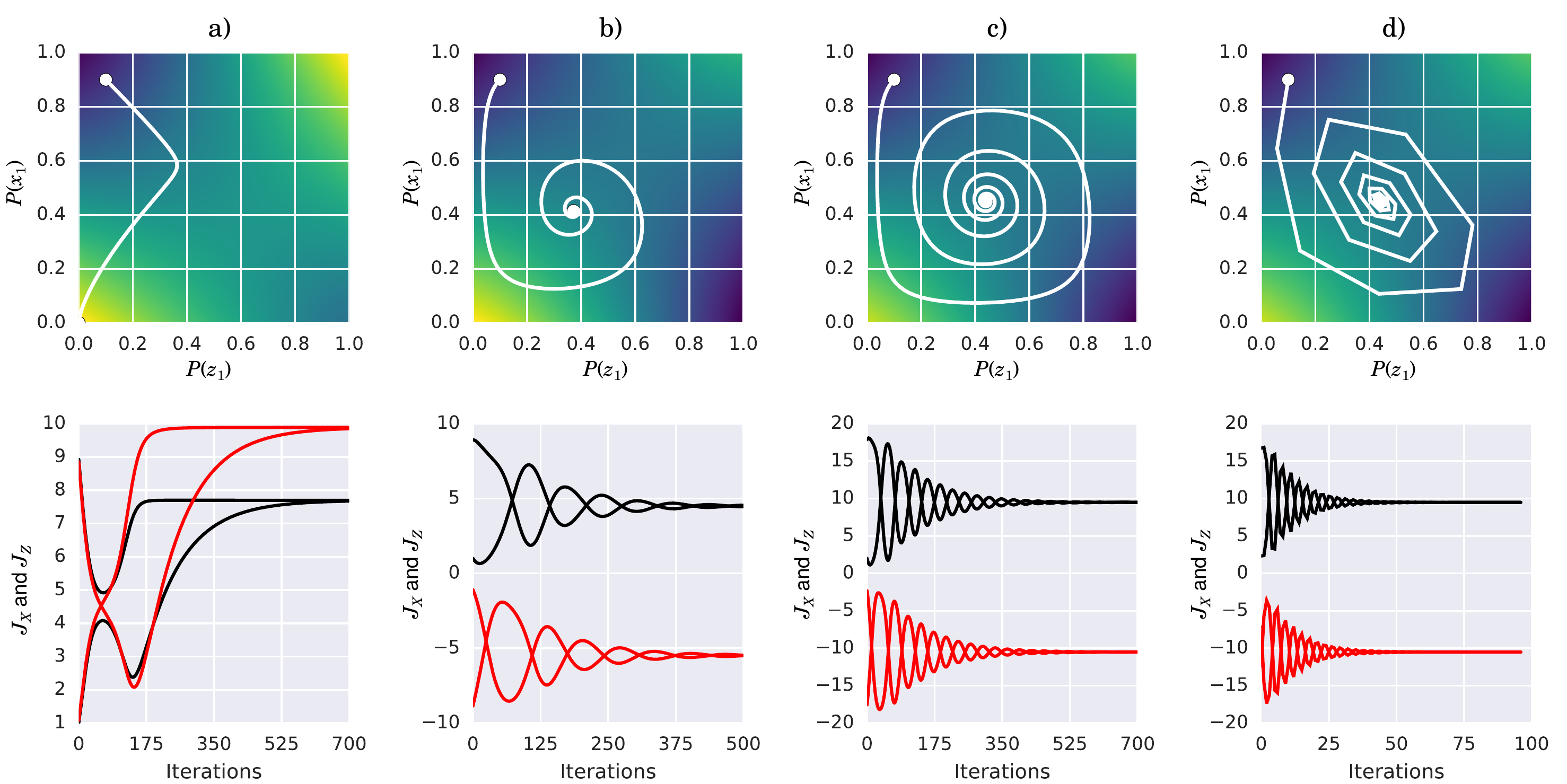}
\par\end{centering}

\caption{\emph{Learning dynamics for computing equilibria.} The plots show four 
learning dynamics for a utility matrix $U=I_{2\times2}$ and prior strategies 
$Q(X)=(0.9,0.1)$ and $Q(Z)=(0.1,0.9)$. To generate the dynamics, the following 
inverse temperatures and learning rates were chosen: a) $\alpha=\beta=10$
and $\eta=0.01$; b)~$\alpha=10$, $\beta=-10$ and $\eta=0.01$;
c) $\alpha=20$, $\beta=-20$ and $\eta=0.01$; and d) $\alpha=20$,
$\beta=-20$ and a larger learning rate $\eta=0.1$. Top row: Evolution
of the strategies. The prior strategies are located in the top-left
corner. The objective function is color-coded and ranges from 0 (dark
blue) to 1.2 (yellow). Bottom row: Evolution of the per-action net~payoffs
$J_{X}$ and $J_{Z}$ defined in equation (\ref{eq:net-payoffs})
illustrating the indifference principle. The black and red curves
correspond to the agent's and the environment's actions 
respectively.\label{fig:learning-dynamics}}
\end{figure}

Now we derive an algorithm for computing the equilibrium strategies
for the agent and the environment. It is well-known that using standard 
gradient descent with competing losses is difficult \citep{Mescheder2017, 
Balduzzi2018}, and indeed a straight-forward gradient-ascent/descent
method on the objective (\ref{eq:objective}) turns out to be numerically
brittle, especially for values of $\alpha$ and $\beta$ near zero.
Rather, we let the strategies follow a smoothed dynamics on the exponential
manifold until reaching convergence. Equation~(\ref{eq:opt-player})
shows that the log-probabilities of the best-response strategies are:
\begin{align*}
\log P(x) & \overset{+}{=}\log Q(x)+\alpha\sum_{z}P(z)U(x,z)\\
\log P(z) & \overset{+}{=}\log Q(z)+\beta\sum_{x}P(x)U(x,z)
\end{align*}
where $\overset{+}{=}$ denotes equality up to a constant. This suggests
the following iterative algorithm. Starting from $L_{0}(x)=\log Q(x)$
and $L_{0}(z)=\log Q(z)$, one can iterate the following four equations
for time steps $t=0,1,2,\ldots$
\begin{alignat}{1}
L_{t+1}(x) & =(1-\eta_{t})\cdot L_{t}(x)+\eta_{t}\cdot\biggl(\log Q(x)+\alpha\sum_{z}P_{t}(z)U(x,z)\biggr)\\
P_{t+1}(x) & =\frac{\exp L_{t+1}(x)}{\sum_{\tilde{x}}\exp L_{t+1}(\tilde{x})}\\
L_{t+1}(z) & =(1-\eta_{t})\cdot L_{t}(z)+\eta_{t}\cdot\biggl(\log Q(z)+\beta\sum_{x}P_{t+1}(x)U(x,z)\biggr)\\
P_{t+1}(z) & =\frac{\exp L_{t+1}(z)}{\sum_{\tilde{z}}\exp L_{t+1}(\tilde{z})}
\end{alignat}
Here, the learning rate $\eta_{t}>0$ can be chosen constant but sufficiently
small to achieve a good approximation; alternatively, one can use
an annealing schedule that conform to the Robbins-Monro conditions 
$\sum_{t}\eta_{t}\rightarrow\infty$ and $\sum_{t}\eta_{t}^{2}<\infty$. 
Figure~(\ref{fig:learning-dynamics}) shows four example simulations of the 
learning dynamics.

\section{Experiments\label{sec:experiments}}

\subsection{Bernoulli Bandit}

We now return to the two-armed bandits discussed in the introduction 
(Figure~\ref{fig:bernoulli-mods}) and explain how these were modeled.

Assuming that the agent plays the rows (i.e.~arms) and the bandit/environment 
the columns (i.e.~reward placement), the utility matrix was chosen as
\[
  U 
  = I_{2\times2}
  = \left[\begin{matrix}
      1 & 0\\
      0 & 1
    \end{matrix}\right]. 
\]
This reflects the fact that there is always one and only one reward.

We did not want the agent to play an equilibrium strategy, but rather 
investigate each bandit's reaction to a uniform strategy and two pure 
strategies: that is, $Q(X)=[0.5, 0.5]^{T}$, $Q(X)=[1, 0]^{T}$, and $Q(X)=[0, 
1]^{T}$ respectively. Then, choosing $\alpha=0$ implies that the agent's 
posterior strategy stays fixed, i.e.\ $P(X)=Q(X)$.

For the bandits, we fixed a common prior strategy $Q(Z)=[0.4,0.6]^{T}$, 
that is, slightly biased toward the second arm. Obviously, appropriate inverse
temperatures were chosen to model the indifferent, friendly, adversarial, 
and very adversarial bandit: 

\begin{center}
\begin{tabular}{clr}
 \toprule
 Bandit & Type & $\beta$ \\ \midrule
 A & Indifferent/Stochastic & $0$ \\
 B & Friendly & $1$ \\
 C & Adversarial & $-1$ \\
 D & Very Adversarial & $-2$ \\ \bottomrule
\end{tabular}
\end{center}

We then computed the equilibrium strategies for each combination, which in 
this case (due to \ $\alpha=0$) reduces to computing the bandits' 
best-response functions for each one of the three agent strategies. Once the 
posterior distributions~$P(Z)$ were calculated, we simulated each one and 
collected the empirical rewards which are summarized in 
Figure~\ref{fig:bernoulli-mods}.

\subsection{Gaussian Bandit, and Dependence on Variance}

In a stochastic bandit, the optimal strategy is to deterministically
pick the arm with the largest expected payoff. However, in adversarial
and friendly bandits, the optimal strategy can depend on the higher-order
moments of the reward distribution, as was shown previously in 
\citep{OrtegaKimLee2015}. The aim of this experiment is to reproduce these 
results, showing the dependence of the optimal strategy on \emph{both the mean 
and variance} of the payoff distribution. 

\paragraph{Setup.} To do so, we considered a four-armed bandit with payoffs 
that are distributed according to (truncated and discretized) Gaussian 
distributions. To investigate the interplay between mean and the variance, we 
chose four Gaussians with:
\begin{itemize}
 \item \emph{increasing means}, where the means $\mu$ are uniformly spaced 
between -0.2 and 0.2;
 \item and \emph{decreasing variances}, where the standard deviations $\sigma$ 
are uniformly spaced between 1 and 2.
\end{itemize}
Thus, the arm with the largest mean is the most precise. Clearly, if the 
bandit is stochastic, arms are ranked according to their mean payoffs 
irrespective of their variances. 

We then performed a sweep through the bandit's inverse temperature
$\beta$, starting from an adversarial ($\beta=-3$) and ending in
a friendly bandit ($\beta=+3$), computing the equilibrium strategies
for both players along the way. Throughout the sweep, the agent's inverse 
temperature was kept constant at $\alpha=30$, modeling a highly rational agent.

\begin{figure}[h]
\begin{centering}
\includegraphics[width=1\textwidth]{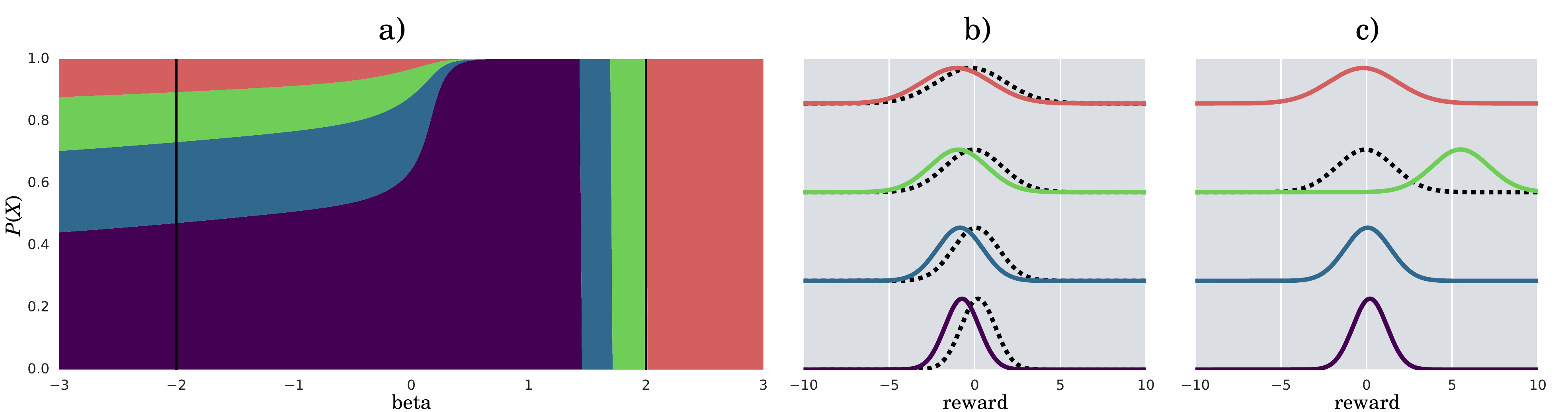}
\par\end{centering}
\caption{\emph{Optimal strategy as a function of $\beta$}. a) Stack plot showing
the action probabilities of the agent's optimal action ($\alpha=30$)
as a function of the inverse temperature parameter of the bandit.
Panels~b \&~c show the payoff distributions of the four arms for
an adversarial ($\beta=-2$) and friendly bandit ($\beta=2$). The
dotted lines represent the prior distributions, and the colored
lines the environment's posterior distributions, color-coded to match
the actions in (a).\label{fig:gaussian-arms}}
\end{figure}

\paragraph{Results.} The results are shown in Figure~\ref{fig:gaussian-arms}.
As expected, for values close to $\beta\approx0$ the agent's optimal
strategy consists in (mostly) pulling the arm with the largest expected
reward. In contrast, adversarial bandits ($\beta<0$) attempt to diminish
the payoffs of the agent's preferred arms, thus forcing it to adopt
a mixed strategy. In the friendly case ($\beta>0$), the agent's strategy
becomes deterministic. Interestingly though, the optimal arm switches
to those with higher variance as $\beta$ increases (e.g.~see Figure~\ref{fig:gaussian-arms}c).
This is because Gaussians with larger variance are ``cheaper to shift''; 
in other words, the rate of growth of the KL-divergence per unit of
translation is lower. Thus, arms that were suboptimal for $\beta=0$ can become 
optimal for larger $\beta$ values if their variances are large. We believe that 
these ``phase transitions'' are related to the ones previously observed under 
information-constraints \citep{TishbySlonim2001, Chechik2005, 
GeneweinLeibfriedGrauMoyaBraun2015}.

\subsection{Linear Classifier}

The purpose of our last experiment is to illustrate the non-trivial 
interactions that may arise between a classifier and a reactive data source, be 
it friendly or adversarial. Specifically, we designed a simple 2-D linear 
classification example in which the agent chooses the parameters of the 
classifier and the environment picks the binary data labels.

\paragraph{Method.} We considered hard classifiers
of the form $y=\sigma(\mathbf{w}^{T}\mathbf{x}-\mathbf{b})$ where
$\mathbf{x}$ and $y$ are the input and the class label, $\mathbf{w}$
and $\mathbf{b}$ are the weight and the bias vectors, and $\sigma$
is a hard sigmoid defined by $\sigma(u)=1$ if $u\geq0$ and $\sigma(u)=-1$
otherwise. To simplify our analysis, we chose a set of 25 data points
(i.e.~the inputs) placed on a $5 \times 5$ grid spread uniformly in 
$[-1,1]^{2}$. Furthermore, we discretized the parameter space so that 
$\mathbf{w}$ and $\mathbf{b}$ have both 25 settings that are uniform in 
$[-1;1]^{2}$ (that is, just as the input locations), yielding a total of 
$25^{2}=625$ possible parameter combinations 
$[\mathbf{w},\mathbf{b}]\in\Theta$. 

The agent's task consisted in choosing a strategy to set those parameters. 
However, unlike a typical classification task, here the agent could choose a 
distribution $P(X = [\mathbf{w},\mathbf{b}])$ over $\Theta$ if deemed 
necessary. Similarly, the environment picked the data labels \emph{indirectly} 
by choosing the parameters of an optimal classifier. Specifically, the 
environment could pick a distribution~$P(Z = [\mathbf{w},\mathbf{b}])$ over 
$\Theta$, which in turn induced (stochastic) labels on the data set.

\begin{figure}[h]
\centering
\includegraphics[width=0.4\textwidth]{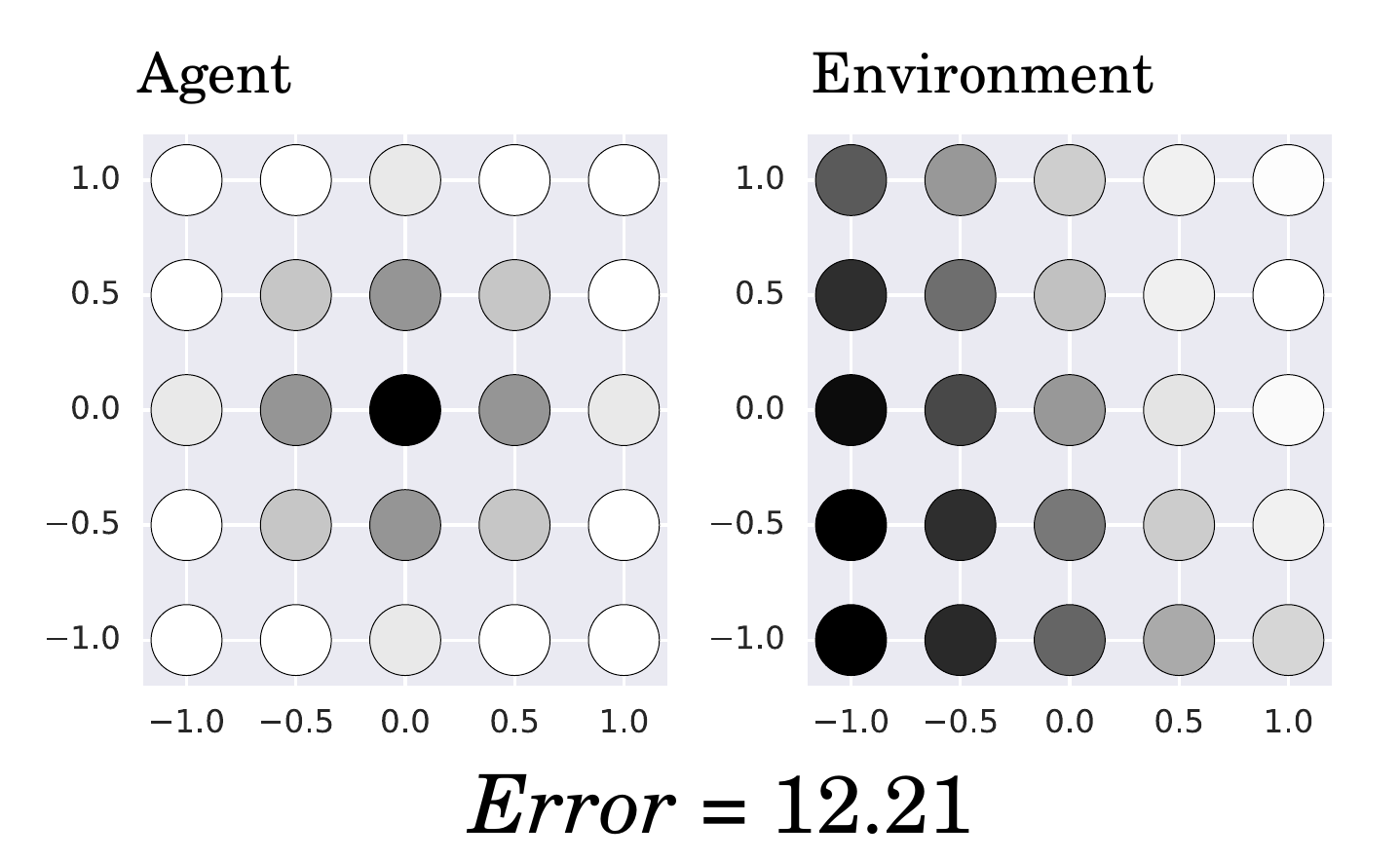}
\caption{\emph{Prior distributions over labels}. The 
diagram shows the (stochastic) labels obtained by 
marginalizing over the linear classifiers chosen by 
the agent (left) and the environment (right).}\label{fig:linear-start}
\end{figure}

The utility and the prior distributions were chosen as follows. The
utility function $U:\Theta\times\Theta\rightarrow\mathbb{\mathbb{N}}$
mapped each classifier-label pair~$(x,z)$ into the number of correctly 
classified data points. The prior distribution of the agent~$Q(X)$ was uniform 
over $\Theta$, reflecting initial ignorance. For the environment we chose a 
prior with a strong bias toward label assignments that are compatible with 
$z^\ast \in \Theta$, where $z^{\ast}=(\mathbf{w}^{\ast},\mathbf{b^{\ast}})$, 
$\mathbf{w}^{\ast}=[-1,-0.5]^{T}$, and~$\mathbf{b}^{\ast}=[-0.5,0.5]^{T}$. 
Specifically, for each label assignment $z \in \Theta$, its prior 
probability $Q(z)$ was proportional to $U(z,z^{\ast})$, the number of 
data points that a model based on $z$ would correctly classify when the true 
labels are given by $z^\ast$ instead. Figure~\ref{fig:linear-start} shows 
the stochastic labels obtained by marginalizing over the prior strategies of 
the agent and the environment respectively. Notice that although
each individual classifier is linear, their mixture is not. 

\paragraph{Results.} 
Starting from the above prior, we then calculated various friendly and 
adversarial deviations. First we improved the agent's best-response strategy
by setting the inverse temperature to $\alpha=30$ and keeping
the environment's strategy fixed ($\beta=0$). As a result, we obtained
a crisper (i.e.~less stochastic) classifier that reduced the mistakes
for the data points that are more distant from the decision boundary.
We will use these posterior strategies as the prior for our subsequent tests.

\begin{center}
\includegraphics[width=0.7\textwidth]{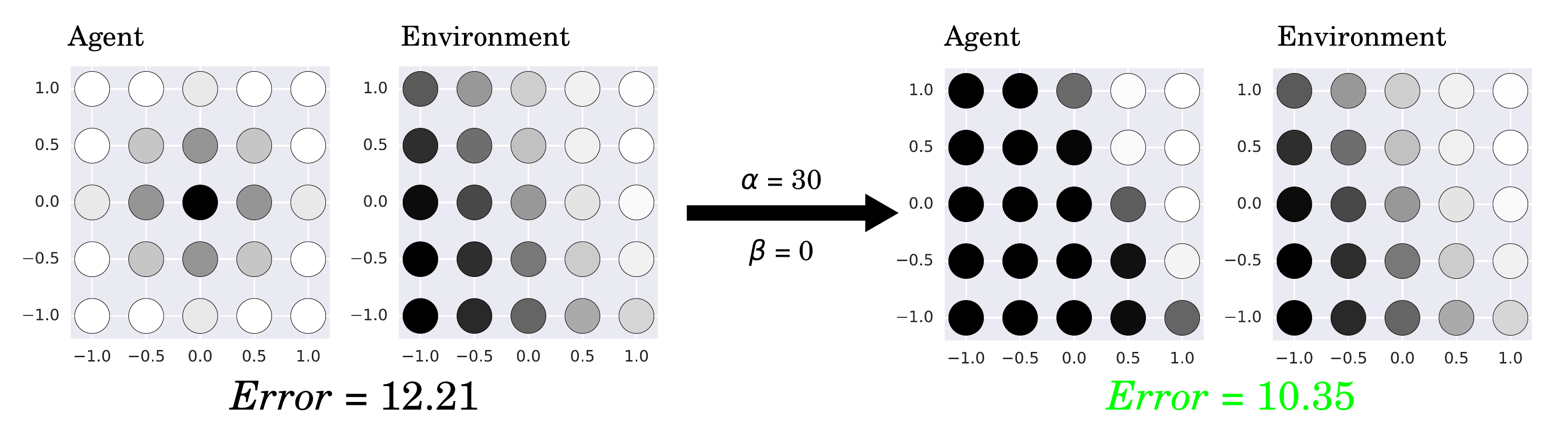}
\end{center}

Next we generated two adversarial modifications
of the environment ($\beta=-0.1$ and $\beta=-1$) while keeping the
agent's strategy fixed, simulating an attack on a pre-trained 
classifier. The case $\beta=-0.1$ shows that a slightly adversarial environment
attempts to increase the classification error by ``shifting'' the
members of the second class (white) towards the agent's decision boundary.
However, a very adversarial environment (case $\beta=-1$) will simply
flip the labels, nearly maximizing the expected classification error.

\begin{center}
\includegraphics[width=0.7\textwidth]{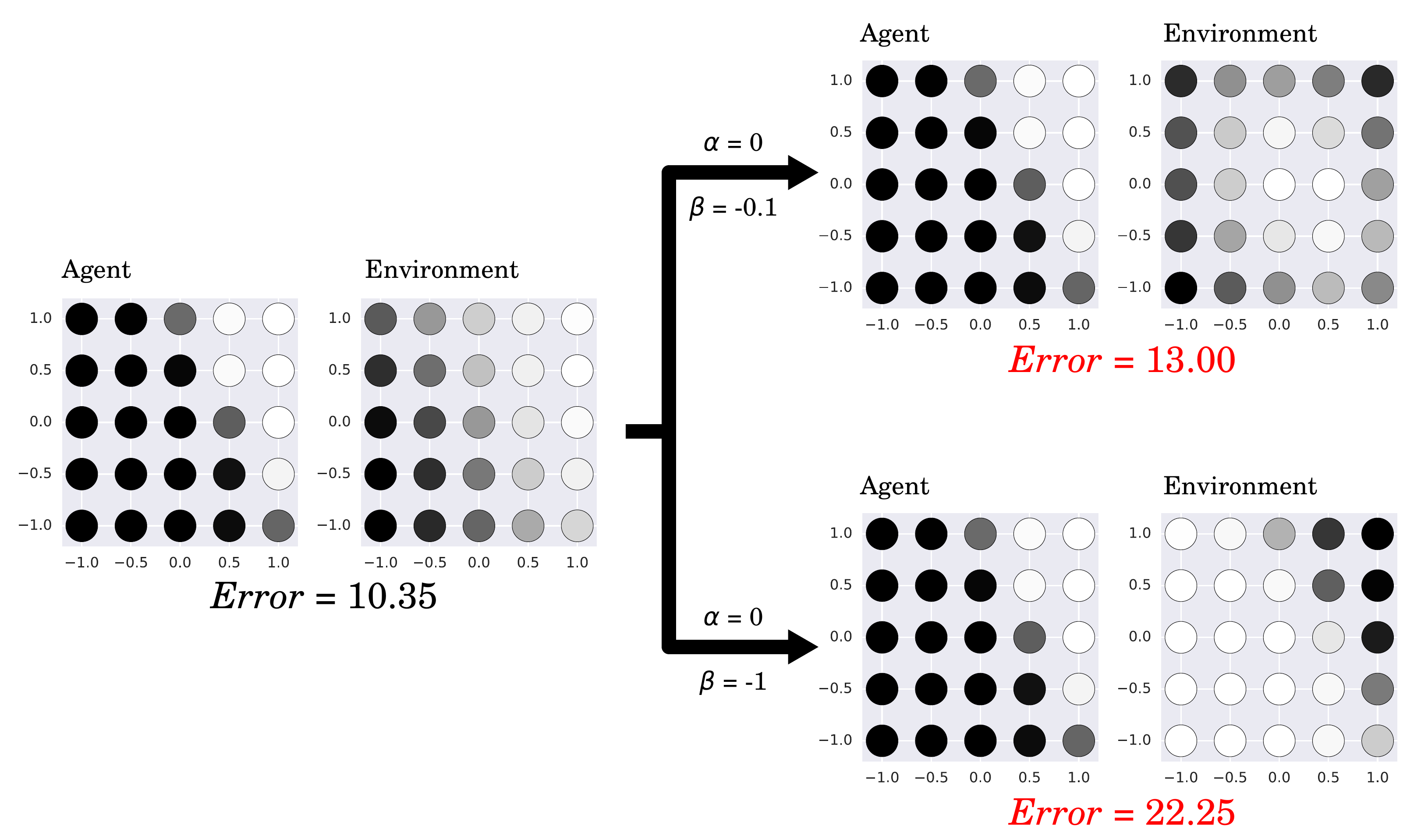}
\end{center}

If instead we pair a reactive agent ($\alpha=10$)
with the previous adversarial environment ($\beta=-1$), we see that
the agent significantly improves his performance by slightly randomizing
his classifier, thereby thwarting the environment's attempt to fully
flip the labels. Finally, we paired a reactive agent
($\alpha=10$) with a friendly environment ($\beta=1$). As a result,
both players cooperated by significantly aligning and sharpening their
choices, with the agent picking a crisp decision boundary that nearly
matched all the labels chosen by the environment.

\begin{center}
\includegraphics[width=0.7\textwidth]{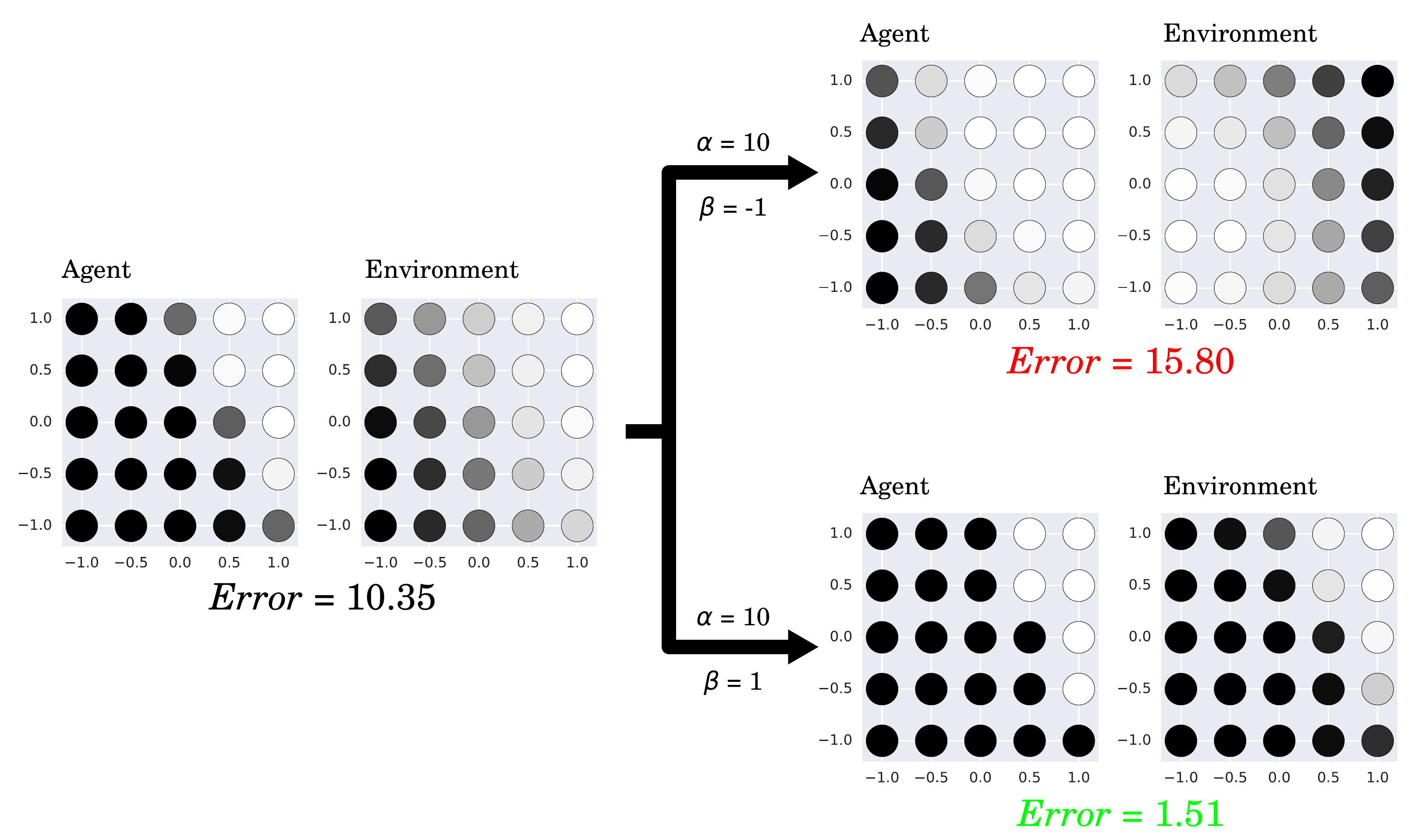}
\end{center}

\section{Discussion and Conclusions}

\subsection{Information Channel}\label{sec:information-channel}

The objective function~\eqref{eq:objective} can be tied more closely to the 
idea of an information channel that allows the agent and the 
environment to anticipate each other's strategy (Section~\ref{sec:rationale}).

Let $D$ be a random variable that encapsulates the totality of the information 
that informs the decisions of the agent and the environment. Identify 
the posteriors $P(X)$ and $P(Z)$ with the mixed strategies that the two 
players adopt after learning about $D$, that is, $P(X) = Q(X|D)$ and $P(Z) 
= Q(Z|D)$ respectively. This corresponds to the graphical model: 
\begin{center}
\includegraphics[width=0.2\textwidth]{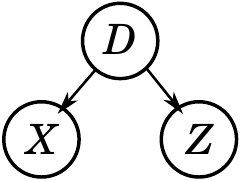}
\end{center}
Assuming $D \sim Q(D)$ and taking the expectation over $D$ of the 
objective function~\eqref{eq:objective}, we obtain
\begin{equation}
  \mathbb{E}_{D}[J]
  = \mathbb{E}_{D,X,Z} \bigl[ U(X, Z) \bigr]
  - \frac{1}{\alpha} I(X; D)
  - \frac{1}{\beta} I(Z; D),
\end{equation}
where $I(X; D)$ and $I(Z; D)$ are mutual information terms. These terms 
quantify the capacity of the information channel between the background 
information~$D$ and the strategies. 

With this connection, we can draw two conclusions. First, The baseline 
strategies $Q(X)$ and $Q(Z)$ are the strategies that result when the two players 
do not observe the background information, since
\[
  Q(x) = \sum_d Q(d) Q(x|d)
  \qquad\text{and}\qquad
  Q(z) = \sum_d Q(d) Q(z|d),
\]
that is, the players play one of their strategies according to their base 
rates, effectively averaging over them. Second, the objective 
function~\eqref{eq:objective} controls, via the inverse temperature parameters, 
the amount of information about $D$ that the agent and environment use to choose 
their strategy.

\comment{
Thus, the objective 
function allows for a precise control of the information 

The notion of friendly and adversarial introduced here can be related to the It 
is important to point that the notion of adversarial used in 
computational learning theory (see for instance \citep{BubeckCesaBianchi2012}) 
is compatible with the zero-sum characterization. which is especially popular 
in the bandit literature () It is important to point out that learning theory 
uses a different notion of adversarial. It is important to point out that the 
learning theory

In addition, the 
definition of \emph{adversarial} differs
from the one proposed here: it is defined as an \emph{arbitrary} behavior,
i.e.~not governed by any probability law~\citep{AuerEtAl2002}.

The work most related to ours is \citep{OrtegaKimLee2015}
in the context of multi-armed bandits. 
}

\subsection{Relation to previous work}

This work builds on a number of previous ideas. The characterization 
of friendly and adversarial from Section~\ref{sec:rationale} is a 
direct adaptation of the Gaussian case introduced in~\citep{OrtegaKimLee2015} to 
the case of discrete strategy sets. Therein, the authors presented
a model of multi-armed bandits for the special case of Gaussian-distributed
rewards in which the bandit can react to the strategy of the agent
in a friendly or adversarial way. They furthermore used this model
to derive the agent's optimal policy and a Thompson sampling algorithm
to infer the environment's inverse temperature parameter from experience.
Our work can be thought as a adaptation of their model to normal-form
games.

In turn, the formalization of friendly and adversarial behavior through 
information-constraints was suggested in \citep{Broek2010} (in the context of 
risk-sensitivity) and in \citep{OrtegaBraun2011, OrtegaBraun2013} (in sequential 
decision-making). Information-constraints have also been used in game theory: 
in particular, Quantal Response Equilibria feature reactive 
strategies that are bounded rational \citep{McKelveyPalfrey1995, 
WolpertHarreOlbrichBertschingerJost2012}, which relates to our case when the 
inverse temperatures are positive. 

Our work was also inspired by existing work in the bandit literature; in 
particular, by bandit algorithms that achieve near-optimal 
performance in both stochastic and adversarial
bandits \citep{BubeckSlivkins2012, SeldinSilvkins2014, AuerChaoKai2016}.
Notice however, that these algorithms do neither include the 
friendly/cooperative case nor distinguish between different degrees of 
attitudes.

In multiagent learning there is work that considers how to act in different 
scenarios with varying, \emph{discrete degrees} of adversarial behaviour (see 
for instance \citep{Littman2001, Powers2005, Greenwald2003, Crandall2011}). In 
particular, a Bayesian approach has been tested to learn against a 
given class of opponents in stochastic games \citep{Hernandez2017}.

\subsection{Learning the attitude of an environment}

In this work we have centered our attention on formalizing friendly
and adversarial behavior by characterizing the \emph{statistics} of such
an environment in a one-shot game given three components: 
\begin{description}
\item[a)] the strategy of the agent; 
\item[b)] the prior strategy of the environment; 
\item[c)] and an inverse temperature parameter. 
\end{description}
This model can be used within a learning algorithm to detect the 
environment's friendly or adversarial attitude from past interactions akin 
to \citep{Littman2001}. 
However, since this sacrifices the one-shot setup, additional assumptions 
(e.g.\ of stationarity) need to be made to accommodate our definition.

For instance, in a Bayesian setup, the model can be used as a likelihood 
function~$P(z|\beta,\theta,\pi)$
of the inverse temperature~$\beta$ and the parameters~$\theta$
of the environment's prior distribution~$P(z|\theta)$ given the
parameters~$\pi$ of the (private) strategy $P(x|\pi)$ used
by the agent. If combined with a suitable prior over~$\beta$ and~$\theta$,
one can e.g.~use Thompson sampling \citep{Thompson1933,ChappelleLi2011}
to implement an adaptive strategy that in round $t+1$ plays the best
response $P(x|\pi_{t+1})$ for simulated parameters~$\beta'_{t+1}$
and~$\theta'_{t+1}$ drawn from the posterior~$P(\beta,\theta|\pi_{1:t},z_{1:t})$.
This is the method adopted in \citep{OrtegaKimLee2015}.

Alternatively, another way of detecting whether the environment is
reactive is by estimating the mutual information $I(\pi;z|x)$ between
the agent's strategy parameter~$\pi$ and the environment's action~$z$
given the player's action. This is because, for a non-reactive environment,
the agent's action $x$ forms a Markov blanket for the environment's
response~$z$ and hence $I(\pi;z|x)=0$; whereas if $I(\pi;z|x)>0$,
then it must be that the environment can ``spy'' on the agent's private
policy.

\subsection{Final thoughts}

We have presented an information-theoretic definition of behavioral
attitudes such as friendly, indifferent, and adversarial and shown
how to derive optimal strategies for these attitudes. These results
can serve as a general conceptual basis for the design of specialized
detection mechanisms and more robust strategies. 

Two extensions are worth pointing out. The first is the extension
of the model to extensive-form games to represent sequential interactions.
This will require formulating novel backward induction procedures
involving subgame equilibria \citep{OsborneRubinstein1994}, perhaps similar to 
\citep{Ling2014}. The second
is the analysis of state-of-the-art machine learning techniques such
as deep neural networks: e.g.~whether randomizing the weights protects
from adversarial examples \citep{SzegedyZarembaSutskeverBrunaErhanGoodfellowFergus2013,GoodfellowShlensSzegedy2014};
and whether friendly examples exist and can be exploited. 

Importantly, we have shown the existence of a continuous range 
of environments that, \emph{if not finely discriminated by the agent, will lead 
to strictly suboptimal strategies}, even in the friendly case.

\subsubsection*{Acknowledgments}
We thank Marc Lanctot, Bernardo Pires, Laurent Orseau, Victoriya Krakovna,  
Jan Leike, Neil Rabinowitz, and David Balduzzi for comments on an earlier 
manuscript.

\appendix

\section{Proofs}\label{sec:proofs}
\paragraph{Proof of Proposition 1. }
The best response-functions are obtained by optimizing the Lagrangian
\[
L:=J-\lambda_{X}\Bigl(\sum_{x}P(x)-1\Bigr)-\lambda_{Z}\Bigl(\sum_{z}P(z)-1\Bigr)
\]
where $\lambda_{X}$ and $\lambda_{Z}$ are the Lagrange multipliers
for the equality constraints enforcing the normalization of $P(X)$
and $P(Z)$ respectively. For $P(Z)$, we fix $P(X)$ and equate the
derivatives to zero:

\[
\frac{\partial L}{\partial 
P(z)}=\sum_{x}P(x)U(x,z)-\frac{1}{\beta}\Bigl(\log\frac{P(z)}{Q(z)}
+1\Bigr)+\lambda_{Z}\overset{!}{=}0.
\]
Solving for $P(z)$ yields
\begin{equation}
P(z)=Q(z)\exp\Bigl\{\beta\sum_{x}P(x)U(x,z)+\beta\lambda_{Z}-1\Bigr\}
\label{eq:unnorm-solution}
\end{equation}

Since $\sum_{z}P(z)=1$, it must be that the Lagrange multiplier $\lambda_{Z}$
is equal to
\[
\lambda_{Z}=-\frac{1}{\beta}\log\sum_{z}Q(z)\exp\Bigl\{\beta\sum_{z}P(x)U(x,
z)-1\Bigr\},
\]
which, when substituted back into (\ref{eq:unnorm-solution}) gives
the best-response function of the claim. The argument for $P(X)$
proceeds analogously. $\blacksquare$

\paragraph{Proof of Proposition 2.}
The combined best-response function is a continuous
map from a compact set into itself. It follows therefore from Brouwer's
fixed-point theorem that it has a fixed-point. $\blacksquare$

\paragraph{Proof of Proposition 3.}
We first multiply (\ref{eq:objective}) by $\alpha$
and then see that, for any fixed $P(Z)$, the agent's best response
$P(X)$ is also the maximizer (that is, irrespective of the sign of 
$\alpha$) of the objective function 
\begin{equation}
\sum_{x}P(x)\Bigl\{\alpha\sum_{z}P(z)U(x,z)-\log\frac{P(x)}{Q(x)}\Bigr\}=\sum_{x
}P(x)J_{X}(x)\label{eq:indiff-1}
\end{equation}
which ignores the terms that do not depend on $P(X)$. $J_{X}$ is
a continuous and monotonically decreasing function in~$P(x)$. Assume
by contradiction that there are actions~$x_{1},x_{2}$ such that
$J_{X}(x_{1})<J_{X}(x_{2})$. Then one can always improve the objective
function by transferring a sufficiently small amount of probability mass 
from action $x_{1}$ to action $x_{2}$. However, this contradicts the 
assumption that $P(X)$ is a best-response and thus, 
$J_{X}(x_{1})=J_{X}(x_{2})$. The argument
for $J_{Z}$ proceeds analogously. $\blacksquare$

\bibliographystyle{unsrtnat}
\bibliography{bibliography}

\end{document}